\documentclass{article}

\usepackage{times}
\usepackage{graphicx} 
\usepackage{subfigure} 

\usepackage{natbib}

\usepackage{microtype}

\usepackage{amsmath}
\usepackage{amssymb}

\usepackage{algorithm}
\usepackage{algorithmic}

\usepackage{hyperref}

\usepackage[accepted]{icml2016} 

\icmltitlerunning{Inference Networks for Sequential Monte Carlo in Graphical Models}

\makeatletter
\newcommand*{\indep}{%
  \mathbin{%
    \mathpalette{\@indep}{}%
  }%
}
\newcommand*{\nindep}{%
  \mathbin{
    \mathpalette{\@indep}{\not}
  }%
}
\newcommand*{\@indep}[2]{%
  \sbox0{$#1\perp\m@th$}
  \sbox2{$#1=$}
  \sbox4{$#1\vcenter{}$}
  \rlap{\copy0}
  \dimen@=\dimexpr\ht2-\ht4-.2pt\relax
  \kern\dimen@
  {#2}%
  \kern\dimen@
  \copy0 
} 
\makeatother

\usepackage{tikz}
\usetikzlibrary{bayesnet}

\usepackage{units}

\newcommand{\Normal}{\mathcal{N}}

\newcommand{\E}{\mathbb{E}}
\newcommand{\params}{\vartheta}

\newcommand{\Y}{\mathcal{Y}}

\newcommand{\x}{\+x}

\newcommand{\xn}{\x_{1:n}}

\newcommand{\xN}{\x_{1:N}}

\newcommand{\Mb}{\textsc{mb}}
\newcommand{\Pa}{\textsc{pa}}
\newcommand{\PaH}{\widetilde{\textsc{pa}}}

\renewcommand{\phi}{\varphi}

\usepackage{wrapfig}

\tikzstyle{factor} = [rectangle,fill=white,draw=black,inner sep=1pt,
minimum size=20pt, font=\fontsize{10}{10}\selectfont, node distance=1]

\tikzstyle{plate caption} = [caption, node distance=0, inner sep=0pt,
below left=0em and 0em of #1.south east] %

\title{Inference Networks for Sequential Monte Carlo in Graphical Models}
\author{Brooks Paige \hspace{2em} Frank Wood \\ University of Oxford \\ {\small \texttt{\{brooks,fwood\}@robots.ox.ac.uk}}}
\date{}

\begin{document} 

\twocolumn[
\icmltitle{Inference Networks for Sequential Monte Carlo in Graphical Models}

\icmlauthor{Brooks Paige}{brooks@robots.ox.ac.uk}
\icmlauthor{Frank Wood}{fwood@robots.ox.ac.uk}
\icmladdress{Department of Engineering Science, University of Oxford}


\vskip 0.3in
]



\begin{abstract}
We introduce a new approach for amortizing inference in directed graphical models
by learning heuristic approximations to stochastic inverses,
designed specifically for use as proposal distributions in sequential Monte Carlo methods.
We describe a procedure for constructing and learning a structured neural network 
which represents an inverse factorization of the graphical model,
resulting in a conditional density estimator that takes as input
particular values of the observed random variables, and returns an
approximation to the distribution of the latent variables.
This recognition model can be learned offline, independent from
any particular dataset, prior to performing inference.
The output of these networks can be used as 
automatically-learned high-quality proposal distributions to
accelerate sequential Monte Carlo 
across a diverse range of problem settings.
\end{abstract}

\section{Introduction}

Recently proposed methods for Bayesian inference based on sequential Monte Carlo \cite{doucet2001sequential} have
shown themselves to provide state-of-the art results in applications far broader
than the traditional use of sequential Monte Carlo (SMC) for filtering in state space models \citep{gordon1993novel,pitt1999filtering},
with diverse application to
factor graphs \citep{naesseth2014sequential}, 
hierarchical Bayesian models \citep{lindsten2014divide},
procedural generative graphics \citep{ritchie2015controlling},
and
general probabilistic programs \citep{wood2014anglican,todeschini2014biips}.
These are accompanied by complementary computational advances,
including memory-efficient implementations \citep{jun2014memory},
and highly-parallel variants \citep{murray2014parallel,paige2014async}.

All these algorithms, however, share the need for specifying a series of {\em proposal distributions},
used to sample candidate values at each stage of the algorithm.
Sequential Monte Carlo methods perform inference progressively, iteratively targeting
a sequence of intermediate distributions which culminates in a final target distribution.
Well-chosen proposal distributions for transitioning from one intermediate target distribution
to the next can lead to sample-efficient inference,
and are necessary for practical application of these methods to difficult inference problems.
Theoretically optimal proposal distributions \citep{doucet2000sequential,cornebise2008} are in general intractable,
thus in practice implementing these algorithms
requires either active (human) work to design an appropriate proposal distribution prior to sampling,
or using an online estimation procedure to approximate the optimal proposal during inference 
(as in e.g.~\citet{van2000unscented} or \citet{cornebise2014} for state-space models).
In many cases, a baseline proposal distribution which simulates from a prior distribution can be used,
analogous to the so-called bootstrap particle filter for inference in state-space models;
however, when confronted with tightly peaked likelihoods (i.e.~highly informative observations), 
proposing from the prior distribution may be arbitrarily statistically inefficient \citep{delmoral2015sequential}.
Furthermore, for some choices of sequences of densities
there is no natural prior distribution, or even it may not be available in closed form.
All in all, the need to design appropriate proposal distributions is a real impediment 
to the automatic application of these SMC methods to new models and problems.

This paper investigates how autoregressive neural network models for modeling probability distributions \citep{bengio1999modeling,uria2013rnade,germain2015made}
can be leveraged to automate the design of model-specific proposal distributions for sequential Monte Carlo.
We propose a method for learning proposal distributions for a given probabilistic generative model offline,
prior to performing inference on any particular dataset.
The learned proposals can then be reused as desired, allowing SMC inference to be performed
quickly and efficiently for the same probabilistic model, but for new data --- that is, for new settings of
the observed random variables --- once we have incurred the up-front cost of learning the proposals.

We thus present this work as an amortized inference procedure in the sense of 
\citet{gershman2014amortized}, 
in that it takes a model as its input
and generates an artifact which then can be leveraged for accelerating future inference tasks.
Such procedures have been considered for other inference methods:
learning idealized Gibbs samplers offline for models in which closed-form full conditionals are not available \citep{stuhlmueller2013learning},
using pre-trained neural networks to inform local MCMC proposal kernels \citep{jampani2015informed,kulkarni2015picture},
and learning messages for new factors for expectation-propagation \citep{heess2013learning}.
In the context of SMC, offline learning of high-quality proposal distributions provides a similar opportunity for amortizing
runtime costs of inference, while simultaneously automating a currently-manual process.

Source code for all experiments (in PyTorch) is available at\\{\small \url{https://github.com/tbrx/compiled-inference}}.

\section{Preliminaries}

A directed graphical model, or Bayesian network \citep{pearl1998bayesian}, defines a joint probability distribution
and conditional independence structure via a directed acyclic graph.
For each $x_i$ in a set of random variables $x_1, \dots, x_N$, the network structure specifies
a conditional density $p_i(x_i|\Pa(x_i))$, where $\Pa(x_i)$ denotes the parent nodes of $x_i$.
Inference tasks in Bayesian networks involve marking certain nodes as observed random variables,
and characterizing the posterior distribution of the remaining latent nodes.
The joint distribution over $N$ latent random variables $\+x$ and $M$ observed random variables $\+y$ 
is defined as
\begin{align}
p(\+x, \+y) \triangleq \prod_{i=1}^N f_i\left(x_i|\Pa(x_i)\right) \prod_{j=1}^M g_j\left(y_j|\Pa(y_j)\right),
\end{align}
where $f_i$ and $g_j$ refer to the probability density or mass functions associated
with the respective latent and observed random variables $x_i, y_j$.
Posterior inference in directed graphical models entails using Bayes' rule to estimate the posterior distribution
of the latent variables $\+x$ given particular observed values $\+y$;
that is, to characterize the target density $\pi(\+x) \equiv p(\+x|\+y)$.
In most models, exact posterior inference is intractable, and one must resort to either variational or finite-sample approximations.

\subsection{Sequential Monte Carlo}

Importance sampling methods approximate expectations with respect to a (presumably intractable) distribution $\pi(\+x)$
by weighting samples drawn from a (presumably simpler) proposal distribution $q(\+x)$.
In graphical models, with $\pi(\+x) \equiv p(\+x|\+y)$,
we define an unnormalized target density $\gamma(\+x) \equiv p(\+x,\+y)$
such that $\pi(\+x) = Z^{-1}\gamma(\+x)$, where the normalizing constant $Z$ is unknown.

The sequential Monte Carlo algorithms we consider \citep{doucet2001sequential} 
for inference on an $N-$dimensional latent space $\xN$
proceed by incrementally importance sampling a weighted set of $K$ particles,
with interspersed resampling steps to direct computation towards more promising regions of the high-dimensional space.
We break the problem of estimating the posterior distribution of $\xN$ into a series of simpler lower-dimensional problems
by constructing an artificial sequence of target densities $\pi_1,\dots,\pi_N$ (and corresponding unnormalized densities $\gamma_1,\dots,\gamma_N$)
defined on increasing subsets $\x_{1:n}$, $n=1,\dots,N$, where the final $\pi_N \equiv \pi$
is the full target posterior of interest.
At each intermediate density, the importance sampling density  $q_{n+1}(x_{n+1}|x_{1:n})$ only needs to adequately approximate
a low-dimensional step from $x_{1:n}$ to $x_{n+1}$.

Procedurally, we initialize at $n=1$ by
sampling $K$ values of $x_1$ from a proposal density $q_1(x_1)$, and 
assigning each of these particles $x_1^k$  an associated 
importance weight
\begin{align}
w_1(x^k_1) &= \frac{\gamma_1(x^k_1)}{q_1(x^k_1)}, & 
W_1^k &= \frac{w_1(x^k_1)}{\sum_{j=1}^K w_1(x_1^j)}.
\end{align}

For each subsequent $n = 2, \dots, N$, we
first resample the particles according to the normalized weights
at $W_{n-1}^k$, preferentially duplicating high-weight particles
and discarding those with low weight.
To do this we draw particle ancestor indices $a^1_{n-1}, \dots, a^K_{n-1}$
from a resampling distribution $r(\cdot|W^1_{n-1},\dots,W^K_{n-1})$
corresponding to any standard resampling scheme \citep{douc05comparisonof}.
We then extend each particle by sampling a value for $\x_n^k$
from the proposal kernel $q_n(x_{n}^k|\cdot)$,
and update the importance weights
\begin{align}
w_{n}(\xn^k) &= \frac{\gamma_n(\xn^k)}{\gamma_{n-1}(\x_{1:n-1}^{a^k_{n-1}})q_n(x^k_n|\x^{a^k_{n-1}}_{1:n-1})},
\label{eq:smc-weight}
\\
W_n^k &= \frac{w_n(\xn^k)}{\sum_{j=1}^K w_n(\xn^j)}.
\end{align}
We can approximate expectations with respect to the target density $\pi(\xN)$ using
the SMC estimator
\begin{align}
\hat \pi(\xN^{1:K}) &= \sum_{k=1}^K W_N^k \delta_{\xN^k}(\xN),
\label{eq:smc-estimator}
\end{align}
where $\delta(\cdot)$ is a Dirac point mass.

\subsection{Target densities and proposal kernels}

The choice of incremental target densities is application-specific;
innovation in SMC algorithms often involves
proposing novel manners for constructing sequences of intermediate distributions.
These incremental densities do not necessarily need to correspond to marginal distributions of full target.
Particularly relevant recent work directed towards improving SMC inference in the same class of models we address
includes the Biips ordering and arrangement algorithm \citep{todeschini2014biips},
the divide and conquer approach \citep{lindsten2014divide},
and heuristics for scoring orderings in general factor graphs \citep{naesseth2014sequential,naesseth2015towards}.
All these methods provide a means for selecting a sequence of intermediate target densities ---
however, given a sequence of targets, one still must supply an appropriate proposal density.

The ideal choice for this proposal in general
is found by proposing directly from the incremental change in densities \citep{doucet2000sequential}, with
\begin{align}
q^{opt}_n(x_n|x_{1:n-1}) &= \frac{\pi_n(x_{1:n})}{\pi_{n-1}(x_{1:n-1})} \propto \frac{\gamma_n(x_{1:n})}{\gamma_{n-1}(x_{1:n-1})}.
\end{align}
Using this proposal, each of the unnormalized weights in Equation~\eqref{eq:smc-weight} are independent of the sampled values of $x_n^k$.
In practice this conditional density is nearly always intractable, and one must resort to approximation.

Adaptive importance sampling methods aim to learn the optimal proposal online during the course of inference,
immediately prior to proposing values for the next target density.
In both in the context of population Monte Carlo (PMC) \cite{cappe2008adaptive} and 
sequential Monte Carlo \citep{cornebise2008,cornebise2014,gu2015neural},
a parametric family $q(\+x|\lambda)$ is proposed, with $\lambda$ is a free parameter, 
and the adaptive algorithms aim to minimize either the reverse Kullback-Leibler (KL) divergence or Chi-squared distance between
the approximating family $q(\+x|\lambda)$ and the optimal proposal density.
This can be optimized via stochastic gradient descent \citep{gu2015neural}, 
or for specific forms of $q$ by online Monte Carlo expectation maximization,
both for population Monte Carlo \citep{cappe2008adaptive}
and in state-space models \citep{cornebise2014}.
Note that this is the reverse of the KL divergence traditionally used in variational inference \citep{jordan1999introduction},
and takes the form of an expectation with respect to the intractable target distribution. 

\subsection{Neural autoregressive distribution estimation}

As a general model class for $q(\+x|\cdot)$,
we adapt recent advances in flexible neural network density estimators,
appropriate for both discrete and continuous high-dimensional data.
We focus particularly on the use of autoregressive neural network 
density estimation models \citep{bengio1999modeling,larochelle2011neural,uria2013rnade,germain2015made}
which model high-dimensional distribution by learning a sequence of one-dimensional conditional distributions; 
that is, learning each product term in
\begin{align}
p(\+x)  = \prod_{n=1}^N p(x_n|x_1,\dots,x_{n-1}),
\end{align}
typically with weight parameter sharing across densities.

We choose to adapt the masked autoencoder for distribution estimation (MADE) model \citep{germain2015made},
which fits an autoregressive model to binary data, with structure inspired by autoencoders.
In its simplest form, a single-layer MADE model described on $N-$dimensional binary data $\+x \in [0,1]^N$ has a hidden layer $\+h(\+x)$ and output $\hat{\+x}$ with
\begin{align}
\+h(\+x) &= \sigma_w(\+b + (\+W \odot \+M_w)\+x) \\
\hat{\+x} &= \sigma_v(\+c + (\+V \odot \+M_v)\+h(\+x)),
\label{eq:MADE}
\end{align}
where $\+b, \+c, \+W, \+V$ are real-valued parameters to be learned, $\odot$ denotes elementwise multiplication,
$\sigma_w, \sigma_v$ are nonlinear functions, and $\+M_w, \+M_v$ are fixed binary masks.
Critically, the construction of the masks is such that computing the network output for each $\hat x_n$
requires only the inputs $x_1, \dots, x_{n-1}$, with the zeros in the masks dropping the connections.
The masks are generated by assigning each unit in each hidden layer a number from $1,\dots,N-1$,
describing which of the dimensions $x_1,\dots,x_{n-1}$ it is permitted to take as input;
output units then are only permitted to take as input hidden nodes numbered lower than their output.

With a logistic function sigmoid as $\sigma_v$, then $\hat{x}_n$ can be interpreted as a probability $p(x_n|x_1,\dots,x_{n-1})$,
and to compute $\hat{x}_n$ one does not need supply any value as input to $\+h(\+x)$
for the dimensions $x_n,\dots,x_N$.
That is, if one follows all connections ``back'' through the network from $\hat{x}_n$ to the input $\+x$, 
one would find only themselves at $x_1,\dots,x_{n-1}$.

\begin{figure*}
\centering

\resizebox{0.85\textwidth}{!}
{
\begin{tikzpicture}

  \node[obs](y){$t_n$};
  \node[obs,above=1.8em of y]  (x) {$z_{n}$}; 
  \node[latent, left=2.2em of x,yshift=0.5em] (w0) {${w_0}$};
  \node[latent, below=.6em of w0] (w1) {${w_1}$};
  \node[latent, below=.6em of w1] (w2) {${w_2}$};

  \edge {x,w0} {y} ; 
  \edge {x,w1} {y} ; 
  \edge {x,w2} {y} ; 
    
  \plate [inner xsep=1em,inner ysep=.4em,yshift=.2em]{xy} {(x)(y)} {\hbox{$N$\hspace{-.5em}}} ;
\end{tikzpicture}
\hspace{12em}
\begin{tikzpicture}

  \node[obs](y){$t_n$};
  \node[obs,above=1.8em of y]  (x) {$z_{n}$};
  \node[latent, right=2.2em of x,yshift=1.0em] (w0) {${w_0}$};
  \node[latent, below=.8em of w0,xshift=1.6em] (w1) {${w_1}$};
  \node[latent, below=.8em of w1,xshift=-1.6em] (w2) {${w_2}$};

  \edge{y}{x,w0,w1,w2}
  \edge{x}{w0,w1,w2}
  \edge{w0}{w1,w2}
  \edge{w1}{w2}

  \plate [inner xsep=.8em,inner ysep=.4em, yshift=.2em]{xy} {(x)(y)} {\hbox{$N$\hspace{-.5em}}} ;
\end{tikzpicture}
\hspace{9em}
%
%
%
\begin{tikzpicture}

  \node[obs](y){$t_n$};
  \node[obs,above=1.6em of y]  (x) {$z_{n}$};

  \node[latent, right=4.5em of x,yshift=1.0em] (w0) {${w_0}$};
  \node[latent, below=.6em of w0,xshift=0.6em] (w1) {${w_1}$};
  \node[latent, below=.6em of w1,xshift=-0.6em] (w2) {${w_2}$};
  \node[factor, left=1.2em of w1] (f)  {$\phi_{w}$};
  
  \edge{y}{f}
  \edge{x}{f}
  \edge{f}{w0}
  \edge{f}{w1}
  \edge{f}{w2}
  
  \plate [inner xsep=.8em,inner ysep=.4em, yshift=.2em]{xy} {(x)(y)} {\hbox{$N$\hspace{-.5em}}} ;
\end{tikzpicture}
}
\vspace{-.1em}
\caption{A non-conjugate regression model, as (left) a Bayes net representing a generative model for the data $\{t_n\}$; 
(middle) with dependency structure inverted, as a generative model for the latent variables $w_0,w_1,w_2$;
(right) showing the explicit neural network structure of the learned approximation to the inverse conditional distribution $\tilde p(w_{0:2} | z_{1:N}, t_{1:N})$.
New datasets $\{z_n, t_n \}_{n=1}^N$ can be input directly into the joint density estimator $\phi_{w}$ to estimate the posterior.
Note that the ordering of the latent variables $w_{0:2}$ used in this example is chosen arbitrarily;
any permutation of the latent variables would not change the overall structure of the inverse model.
}
\vspace{-.3em}
\label{fig:regression}
\end{figure*}
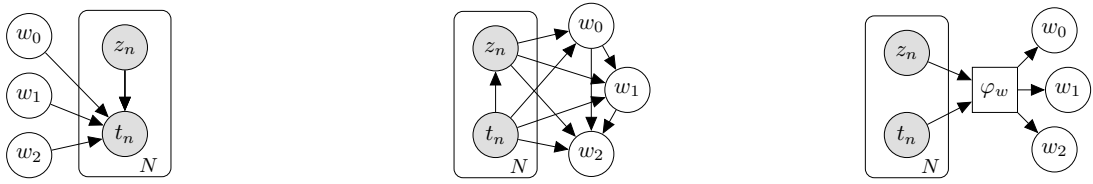

\section{Approach}
\label{sec:approach}

Our approach is two-fold. 
First, given a Bayesian network that acts as a generative model for our observed data $\+y$
given latent variables $\+x$, we construct a new 
Bayesian network which acts as a generative model for our latent $\+x$, given observed data $\+y$.
This network is constructed such that the joint distribution
of the new ``inverse model'', which we will refer to as $\tilde p(\+x,\+y) = \tilde p(\+y)\tilde p(\+x|\+y)$,
preserves the conditional dependence structure in the original model
$p(\+x,\+y) = p(\+x)p(\+y|\+x)$, but has a different factorization \citep{stuhlmueller2013learning}.

Unfortunately, unlike the original forward model, the inverse model has conditional densities which
we do not in general know how to normalize or sample from.
However, were we to know the full conditional density of the inverse model $\tilde p(\+x|\+y)$, then 
we could directly draw samples of $\+x$ given a particular dataset $\+y$.

Thus our second task is to learn approximations for the conditionals $\tilde p(x_i | \PaH(\+x_i))$,
where $\PaH(x_i)$ are parents of $x_i$ in the inverse model.
To do so we employ neural density estimators
and design a procedure to train these ``offline'', in the sense that no real data is required.

\subsection{Defining the inverse model}
\label{sec:inverses}

We begin by constructing an inverse model $\tilde p(\+x,\+y)$
which admits the same distribution over all random variables as $p(\+x,\+y)$, but with a different factorization.
We first note that the directed acyclic graph structure of $p(\+x,\+y)$ 
imposes a partial ordering on all random variables $\+x$ and $\+y$;
we choose any single valid ordering arbitrarily, and define the sequences $x_1, \dots, x_N$ 
and $y_1, \dots, y_M$ 
such that for any $x_i$, $\Pa(x_i) \subseteq \{ x_1, \dots, x_{i-1} \} \cup \{ y_j \}_{j=1}^M$,
and for any $y_j$, $\Pa(y_j) \subseteq \{ y_1, \dots, y_{j-1} \} \cup \{ x_i \}_{i=1}^N$.

Our goal here is to construct as simple as possible a distribution $\tilde p(\+x|\+y)$ whose factorization does not introduce any new conditional independencies not also present in the original generative model.
Consider two extremes: a fully factorized $\tilde p(\+x|\+y) \equiv \prod_{i=1}^N \tilde p(x_i|\+y)$ 
which assumes all $x_i$ are conditionally independent given $\+y$ may be attractive for computational reasons,
but fails to capture all the structure of the posterior;
whereas a fully connected $\tilde p(\+x|\+y) \equiv \prod_{i=1}^N \tilde p(x_i|x_{1:i-1}, \+y)$ is guaranteed to capture all dependencies,
but may be unnecessarily complex.

To define the approximating distribution at each $x_i$, we invert the dependencies on $y_j$,
effectively running the generative model backwards.
Following the heuristic algorithm of \citet{stuhlmueller2013learning},
we do this by literally constructing the dependency graph in reverse.
Ordering the random variables $y_M,\dots,y_1,x_N,\dots,x_1$,
we define a new parent set $\PaH(x_i)$ for each $x_i$ in the transformed model,
with $\PaH(x_i) \subseteq \{ x_{i+1}, \dots, x_N, y_1,\dots,y_M \}$.
Define the Markov blanket $\Mb(x_i)$ to be the set of all random variables which
share a factor with $x_i$; that is, the union of the parents of $x_i$, the children of $x_i$, and the parents of the children of $x_i$.
Then defining the parent sets in the transformed model as
\begin{align*}
\PaH(x_i) &= \Mb(x_i) \cap \{ x_{i+1}, \dots, x_N, y_1,\dots,y_M \} 
\\
\PaH(y_j) &= \Mb(y_j) \cap \{ y_{j+1}, \dots, y_M \}
\end{align*}
yields a model with the same local dependency structure as the original model $p(\+x,\+y)$;
however, now the sequence is reversed such that the observed values are inputs (i.e., $\PaH(y_j) \cap \+x = \emptyset$).
The sequence under the new model, which we will refer to as $\tilde p(\+x,\+y)$, factorizes
naturally as $\tilde p(\+x,\+y) = \tilde p(\+x|\+y)\tilde p(\+y)$;
particularly important to us is the factorization of the conditional density $\tilde p(\+x|\+y) = \prod_{i=1}^N \tilde p(x_i|\PaH(x_i))$.

This algorithm produces inverse graph structures which despite not being fully connected, 
preserve local conditional dependencies in the original graph:

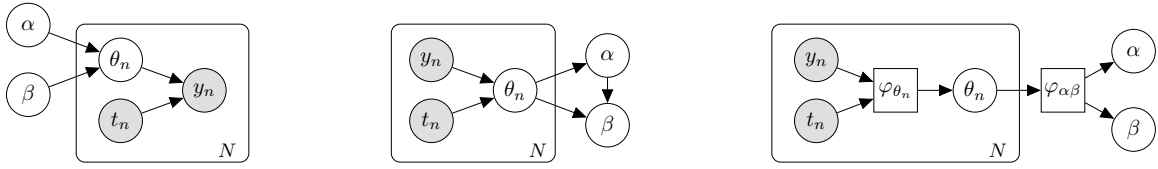
\begin{figure*}[tb]
\centering

\resizebox{0.9\textwidth}{!}{
\begin{tikzpicture}

\node[obs](x){$y_n$};
\node[latent,left=1.8em of x,yshift=1.4em]  (lambda) {$\theta_{n}$};
\node[obs, left=1.8em of x,yshift=-1.4em] (t) {${t_n}$};
\node[latent,left=2.2em of lambda,yshift=1.6em] (alpha) {${\alpha}$};
\node[latent,left=2.2em of lambda,yshift=-1.6em] (beta) {${\beta}$};

\edge {t,lambda} {x} ; 
\edge {alpha,beta} {lambda} ; 
    
\plate [inner xsep=1em,inner ysep=.4em,yshift=.2em]{x} {(x)(t)(lambda)}  {\hbox{$N$\hspace{-.5em}}} ;
\end{tikzpicture}
\hspace{6em}
\begin{tikzpicture}

\node[obs](x){$y_n$};
\node[latent,right=1.8em of x,yshift=-1.4em]  (lambda) {$\theta_{n}$};
\node[obs, left=1.8em of lambda,yshift=-1.4em] (t) {${t_n}$};
\node[latent,right=2.2em of lambda,yshift=1.6em] (alpha) {${\alpha}$};
\node[latent,right=2.2em of lambda,yshift=-1.6em] (beta) {${\beta}$};

\edge {t,x} {lambda} ; 
\edge {lambda} {alpha,beta} ; 
\edge {alpha}{beta};
    
\plate [inner xsep=0.8em,inner ysep=.4em,yshift=.2em]{x} {(x)(t)(lambda)}  {\hbox{$N$\hspace{-.5em}}} ;
\end{tikzpicture}
\hspace{6em}
%
%
%
\begin{tikzpicture}

\node[obs](x){$y_n$};
\node[factor,right=1.6em of x,yshift=-1.4em]  (fn) {$\phi_{\theta_{n}}$};
\node[latent,right=1.6em of fn]  (lambda) {$\theta_{n}$};
\node[obs, left=1.6em of fn,yshift=-1.4em] (t) {${t_n}$};
\node[factor,right=2.0em of lambda] (phi) {${\phi_{\alpha\beta}}$};
\node[latent,right=1.2em of phi,yshift=1.8em] (alpha) {${\alpha}$};
\node[latent,right=1.2em of phi,yshift=-1.8em] (beta) {${\beta}$};

\edge {t,x} {fn} ; 
\edge{fn}{lambda};
\edge {lambda} {phi} ; 
\edge{phi}{alpha,beta};

\plate [inner xsep=1em,inner ysep=.4em,yshift=.2em]{x} {(x)(t)(lambda)} {\hbox{$N$\hspace{-.5em}}} ;
\end{tikzpicture}
}

\caption{A hierarchical Bayesian model. (left) A generative model for the data $\{x_n\}$; 
(middle) with dependency structure inverted;
(right) showing the two distinct joint neural conditional density estimators.
Note in particular the inverse model still partially factorizes across the latent variables.
The learned factor $\phi_{\theta_n}$ is replicated $N$ times in the inverse model, allowing re-use of weights, simplifying training.
}
\label{fig:poisson}
\end{figure*}

{\bf Proposition 1.}
{\it Preservation of local conditional dependence.}
Let $x_A, x_B, x_C$ be latent or observed random variables in $p(\+x)$ with graph structure ${G}$, 
and with each of $x_A, x_B, x_C$ adjacent to at least one of the others under $G$.
Then let $\tilde x_A, \tilde x_B, \tilde x_C$ denote the corresponding random variables in the inverse model $\tilde p(\+x)$ with graph structure $\widetilde{G}$,
constructed via the algorithm above.
If $\tilde x_A$ and $\tilde x_B$ are conditionally independent given $\tilde x_C$ in the inverse model $\widetilde{G}$,
they were also conditionally independent in the original model $G$; that is,
\begin{align}
\tilde x_A \indep \tilde x_B \big | \tilde x_C &\hspace{1em}\Rightarrow\hspace{1em} x_A \indep x_B \big | x_C. \nonumber
\end{align}

{\it Proof.}
Suppose we had a conditional dependence in $G$ which was not preserved in $\widetilde G$, i.e.~with $x_A \nindep x_B \big | x_C$
but $\tilde x_A \indep \tilde x_B \big | \tilde x_C$.
Without loss of generality assume $\tilde x_B$ was added to the inverse graph prior to $\tilde x_A$, i.e.~$x_A \prec x_B$ in $G$.
Note that $x_A \nindep x_B \big | x_C$ can occur either due to a direct dependence between $x_A$ and $x_B$,
or, due to both $x_A, x_B \in \Pa(x_C)$;
in either case, $x_B \in \Mb(x_A)$.
Then when adding $\tilde x_B$ to the inverse graph $\widetilde{G}$
we are guaranteed to have $\tilde x_B \in \PaH(\tilde x_A)$, in which case $\tilde x_A \nindep \tilde x_B$.
$\square$

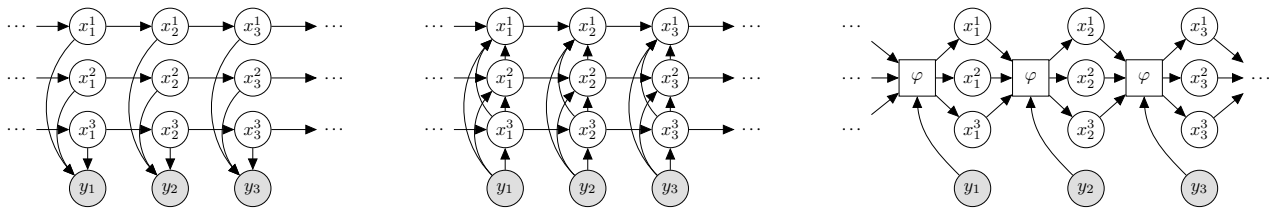
\begin{figure*}
\centering

\resizebox{1.0\textwidth}{!}{
\begin{tikzpicture}

\node[latent](x1){${x}_1^3$};
\node[latent,left=1.8em of x1,draw=none](x0){\dots}; 
\node[latent,right=2.5em of x1](x2){${x}_2^3$};
\node[latent,right=2.5em of x2](x3){${x}_3^3$};
\node[latent,right=2.5em of x3,draw=none](x4){\dots}; 
\node[latent, above=.8em of x0,draw=none](z0){\dots};
\node[latent, above=.8em of x1](z1){$x_1^2$};
\node[latent, above=.8em of x2](z2){$x_2^2$};
\node[latent, above=.8em of x3](z3){$x_3^2$};
\node[latent, above=.8em of x4,draw=none](z4){\dots}; 
\node[latent, above=.8em of z0,draw=none](z10){\dots};
\node[latent, above=.8em of z1](z11){$x_1^1$};
\node[latent, above=.8em of z2](z12){$x_2^1$};
\node[latent, above=.8em of z3](z13){$x_3^1$};
\node[latent, above=.8em of z4,draw=none](z14){\dots}; 
\node[obs, below=1.2em of x1] (y1) {$y_1$};
\node[obs, below=1.2em of x2] (y2) {$y_2$};
\node[obs, below=1.2em of x3] (y3) {$y_3$};

\edge{x1}{x2,y1};
\edge{x2}{x3,y2};
\edge{x3}{x4,y3};
\edge{x0}{x1};
\edge{z0}{z1};
\edge{z1}{z2};
\edge{z2}{z3};
\edge{z3}{z4};
\edge{z10}{z11};
\edge{z11}{z12};
\edge{z12}{z13};
\edge{z13}{z14};
\edge[out=225,in=135]{z1}{y1};
\edge[out=225,in=135]{z2}{y2};
\edge[out=225,in=135]{z3}{y3};
\edge[out=225,in=135]{z11}{y1};
\edge[out=225,in=135]{z12}{y2};
\edge[out=225,in=135]{z13}{y3};
\end{tikzpicture}
\hspace{3em}
\begin{tikzpicture}

\node[latent](x1){${x}_1^3$};
\node[latent,left=1.8em of x1,draw=none](x0){\dots}; 
\node[latent,right=2.5em of x1](x2){${x}_2^3$};
\node[latent,right=2.5em of x2](x3){${x}_3^3$};
\node[latent,right=2.5em of x3,draw=none](x4){\dots}; 
\node[latent, above=.8em of x0,draw=none](z0){\dots};
\node[latent, above=.8em of x1](z1){$x_1^2$};
\node[latent, above=.8em of x2](z2){$x_2^2$};
\node[latent, above=.8em of x3](z3){$x_3^2$};
\node[latent, above=.8em of x4,draw=none](z4){\dots}; 
\node[latent, above=.8em of z0,draw=none](z10){\dots};
\node[latent, above=.8em of z1](z11){$x_1^1$};
\node[latent, above=.8em of z2](z12){$x_2^1$};
\node[latent, above=.8em of z3](z13){$x_3^1$};
\node[latent, above=.8em of z4,draw=none](z14){\dots}; 
\node[obs, below=1.2em of x1] (y1) {$y_1$};
\node[obs, below=1.2em of x2] (y2) {$y_2$};
\node[obs, below=1.2em of x3] (y3) {$y_3$};

\edge{x1}{x2};
\edge{x2}{x3};
\edge{x3}{x4};
\edge{y1}{x1};
\edge{y2}{x2};
\edge{y3}{x3};
\edge{x0}{x1};
\edge{z0}{z1};
\edge{z1}{z2};
\edge{z2}{z3};
\edge{z3}{z4};
\edge{z10}{z11};
\edge{z11}{z12};
\edge{z12}{z13};
\edge{z13}{z14};
\edge[out=135,in=225]{y1}{z1,z11};
\edge[out=135,in=225]{y2}{z2,z12};
\edge[out=135,in=225]{y3}{z3,z13};
\edge[out=135,in=225]{x1}{z11};
\edge[out=135,in=225]{x2}{z12};
\edge[out=135,in=225]{x3}{z13};
\edge{x1}{z1};
\edge{x2}{z2};
\edge{x3}{z3};
\edge{z1}{z11};
\edge{z2}{z12};
\edge{z3}{z13};
\end{tikzpicture}
%
%
\hspace{3em}
\begin{tikzpicture}

\node[latent](x1){${x}_1^3$};
\node[latent,right=4.2em of x1](x2){${x}_2^3$};
\node[latent,right=4.2em of x2](x3){${x}_3^3$};
\node[latent, above=.8em of x1](z1){$x_1^2$};
\node[latent, above=.8em of x2](z2){$x_2^2$};
\node[latent, above=.8em of x3](z3){$x_3^2$};
\node[latent, above=.8em of z1](z11){$x_1^1$};
\node[latent, above=.8em of z2](z12){$x_2^1$};
\node[latent, above=.8em of z3](z13){$x_3^1$};
\node[obs, below=1.2em of x1] (y1) {$y_1$};
\node[obs, below=1.2em of x2] (y2) {$y_2$};
\node[obs, below=1.2em of x3] (y3) {$y_3$};
\node[factor, left=1em of z1](f1){$\phi$};
\node[factor, left=1em of z2](f2){$\phi$};
\node[factor, left=1em of z3](f3){$\phi$};
\node[latent, left=1.5em of f1,draw=none](f0){\dots};
\node[factor, above=0.8em of f0,draw=none](f10){\dots};
\node[factor, below=0.8em of f0,draw=none](f20){\dots};
\node[factor,right=1.4em of z3,draw=none](f4){\dots}; 


\edge{f0,f10,f20}{f1};
\edge{z3,x3,z13}{f4};
\edge{x1,z1,z11}{f2};
\edge{x2,z2,z12}{f3};
\edge{f1}{x1,z1,z11};
\edge{f2}{x2,z2,z12};
\edge{f3}{x3,z3,z13};
\edge[out=135,in=270]{y1}{f1};
\edge[out=135,in=270]{y2}{f2};
\edge[out=135,in=270]{y3}{f3};
\end{tikzpicture}
%
%
}

\caption{Factorial HMM.
(left) The generative model consists $D$ independent Markov models,
with observed data $y_t$ depending on the current state of each latent HMM.
(middle) An inverse model obtained by reversing the order of the generative model at each $t$.
Conditioned on the previous latent states at $t-1$ and the next observation $y_t$, 
all latent states at each $t$ are dependent on one another and must be modeled jointly.
(right) The repeated structure at each $t=1,2,\dots$ means that the same learned conditional density network
can be reused at every $t$.}
\label{fig:factorial-hmm}
\end{figure*}

Examples of generative models and their corresponding inverse models are shown in Figures~\ref{fig:regression}--\ref{fig:factorial-hmm}.
Note that as the topological sort of the nodes in the original generative model is not unique,
neither is the inverse graphical model.

\subsection{Learning a family of approximating densities}

Following \citet{cappe2008adaptive}, 
learning proposals for importance sampling on $\pi(\+x)$ in a single-dataset setting (i.e., with fixed $\+y$)
entails proposing a parametric family $q(\+x|\lambda)$, where $\lambda$ is a free parameter,
and then choosing $\lambda$ to minimize
\begin{align}
D_{KL}(\pi||q_\lambda) &= \int \pi(\+x) \log \left[ \frac{\pi(\+x)}{q(\+x|\lambda)} \right] \mathrm{d}\+x.
\label{eq:fixed-objective}
\end{align}
This KL divergence between the true posterior distribution $\pi(\+x) \equiv p(\+x|\+y)$ and 
proposal distribution $q(\+x|\lambda)$ is also known as the relative entropy criterion,
and is a preferred objective function in situations in which the estimation goal
construct a high-quality weighted sample representation, rather than to minimize the variance of a particular expectation \citep{cornebise2008}.

In an amortized inference setting, instead of learning $\lambda$ explicitly for a fixed value of $\+y$,
we learn a mapping from $\+y$ to $\lambda$.
More explicitly, if $\+y \in \Y$ and $\lambda \in \params$,
then learning a deterministic mapping $\phi : \Y \rightarrow \params$ allows performing approximate 
inference for $p(\+x|\+y)$ with only the computational complexity of evaluating the function $\phi$.
The tradeoff is that the training of $\phi$ itself may be quite involved.

We thus generalize the adaptive importance sampling algorithms by learning a family of distributions $q(\+x|\+y)$,
parameterized by the observed data $\+y$. 
Suppose that $\lambda = \phi(\eta, \+y)$,
where the function $\phi$ is parameterized by a set of upper-level parameters $\eta$.
We would like a choice of $\eta$ which performs well across all datasets $\+y$.
We can frame this as minimizing the expected value of Eq.~\eqref{eq:fixed-objective} under $p(\+y)$,
suggesting an objective function $\mathcal{J}(\eta)$ defined as
\begin{align}
\mathcal{J}(\eta) 
&= \int D_{KL}(\pi||q_\lambda) p(\+y) \mathrm{d}\+y \nonumber \\
&= \int  p(\+y) \hspace{-.2em}\int  p(\+x|\+y) \log \left[ \frac{ p(\+x|\+y)}{q(\+x|\phi(\eta, \+y))} \right] \mathrm{d}\+x \mathrm{d}\+y
\nonumber \\
&= \E_{p(\+x,\+y)} \left[ - \log q(\+x|\phi(\eta, \+y)) \right ] + const
\label{eq:objective}
\end{align}
which has a gradient
\begin{align}
\nabla_\eta 
\mathcal{J}(\eta) 
= \E_{p(\+x,\+y)} \left [ - \nabla_\eta \log q(\+x|\phi(\eta, \+y)) \right ].
\label{eq:gradient}
\end{align}
Notice that these expectations in Equations \eqref{eq:objective} and \eqref{eq:gradient} 
are with respect to the tractable joint distribution $p(\+x,\+y)$.
We can thus fit $\eta$ by stochastic gradient descent,
estimating the expectation of the gradient $\nabla_\eta \mathcal{J}(\eta)$
by sampling synthetic full-data training examples $\{\+x, \+y\}$ from the original model.
This procedure can be performed entirely offline --- we require only to be able to sample from the joint distribution
$p(\+x,\+y)$ to generate candidate data points (effectively providing infinite training data).
In any directed graphical model this can be achieved by ancestral sampling, where in addition to sampling $\+x$
we sample values of the as-yet unobserved variables $\+y$.
Furthermore, we do not need need to be able to compute gradients of our model $p(\+x,\+y)$ itself ---
we only need the gradients of our recognition model $q(\+x|\phi(\eta,\+y))$, allowing use of any differentiable
representation for $q$.
We choose the parametric family $q(\+x|\lambda)$ and the transformation $\phi$ 
such that this inner gradient in Eq.~\eqref{eq:gradient} can be computed easily.

We can now use the conditional independence structure in our inverse model $\tilde p(\+x,\+y)$
to break down $q(\+x|\lambda)$, an approximation of $\tilde p(\+x|\+y)$,
into a product of smaller conditional densities
each approximating $\tilde p(x_i|\PaH(x_i))$.
The full proposal density $q(\+x|\phi(\eta, \+y))$ can be decomposed as
\begin{align}
q(\+x|\phi(\eta, \+y)) = \prod_{i=1}^N q_i(x_i|\phi_i(\eta_i, \PaH(x_i)))
\end{align}
with the gradient similarly decomposing as
\begin{align}
\nabla_{\eta_i} \mathcal{J}(\eta) &= \E_{p(\+x,\+y)} \left [ - \nabla_{\eta_i} \log q_i(x_i|\phi_i(\eta_i, \PaH(x_i))) \right ]. \nonumber
\end{align}
Each of these expectations requires only samples of the random variables in $\{ x_i \} \cup \PaH(x_i)$,
reducing the dimensionality of the joint optimization problem.


\subsection{Joint conditional neural density estimation}

We particularly wish to construct the inverse factorization $\tilde p(\+x|\+y)$ (and our proposal model $q(\cdot)$)
in such a way that we deal naturally with the presence of head-to-head nodes, in which one random variable may have a very large parent set.
This situation is common in machine learning models:
it is quite common to have generative models which factorize in the joint distribution, but have complex dependencies in the posterior;
see for example the model in Figure~\ref{fig:regression}.

We thus choose to treat all such situations in our inverse factorization --- where a sequence of variables $\+x' \subseteq \+x$ are fully dependent on
one another after conditioning on a shared set of parent nodes $\PaH(\+x')$ --- as a single joint conditional density which we 
will approximate with an autoregressive density model.
We extend MADE \citep{germain2015made} to function as a conditional density estimator by
allowing it to take $\PaH(\+x')$ as additional inputs, and constructing the masks such that these additional
inputs are propagated through all hidden layers to all outputs, even for the very first dimension.
As in MADE this can be achieved by labeling the hidden units with integers denoting which input dimensions 
they are allowed to accept.
In contrast to the original MADE, we label hidden units with numbers from $0,\dots,N-1$, 
where hidden units labeled 0 to take as input only the dimensions in $\PaH(\+x')$.
For single-dimensional data, where $N=1$, all hidden units are labeled 0 and all feed forward into the single output $x_1$,
recovering a standard mixture density network \citep{bishop1994mixture}.

To model non-binary data, MADE can be extended 
by altering the output layer network to emit parameters of any univariate probability density function.
We take the same approach by which RNADE \citep{uria2013rnade} modifies the binary autoregressive distribution estimator 
NADE \citep{larochelle2011neural} to handle real-valued data, with an output layer 
that parameterizes a univariate mixture of $D$ Gaussians for each dimension $x_i$ conditioned on its parents. 
The probability of any particular $x_i$ is given by
\begin{align}
q(x_i|\phi_i(\eta_i, \PaH(x_i)))
&=
\sum_{d=1}^D \alpha_{i,d} \mathcal{N}(x_i| \mu_{i,d}, \sigma^2_{i,d}) \nonumber
\end{align}
where $\mathcal{N}(\cdot)$ is the Gaussian probability density.
This requires an output layer with $3\times D$ dimensions,
to predict $D$ each of means $\mu_{i,d}$, standard deviations $\sigma_{i,d}$, weights $\alpha_{i,d}$; to enforce positivity of standard deviations we apply a softplus function
to the raw network outputs, and a softmax function to ensure $\alpha_{i,\cdot}$ is a probability vector.

\subsection{Training the neural network}

Contrary to many standard settings in which one is limited by the amount of data present,
we are armed with a sampler $p(\+x,\+y)$ which allows us to generate effectively infinite training data.
This could be used to sample a ``giant'' synthetic dataset, which we then use for mini-batch training via gradient descent; 
however, then we must decide how large a dataset is required.
Alternatively, we could sample a brand new set of training examples for every mini-batch, never re-using previous samples.

In testing we found that a hybrid training procedure, which samples new synthetic datasets based on performance on
a held-out set of synthetic validation data, appeared more efficient than resampling a new synthetic dataset for each new gradient update.
We perform mini-batch gradient updates on $\eta$ using synthetic training data, while evaluating on the validation set.
If the validation error increases, or after a set maximum number of steps, we draw new sets of both synthetic training and validation data from $p(\+x,\+y)$.

In all experiments we use Adam \citep{kingma2015adam} with the suggested default parameters to update
learning rates online, and use rectified linear activation functions.

\begin{figure*}[tb]
\centering
	\includegraphics[height=1.05in]{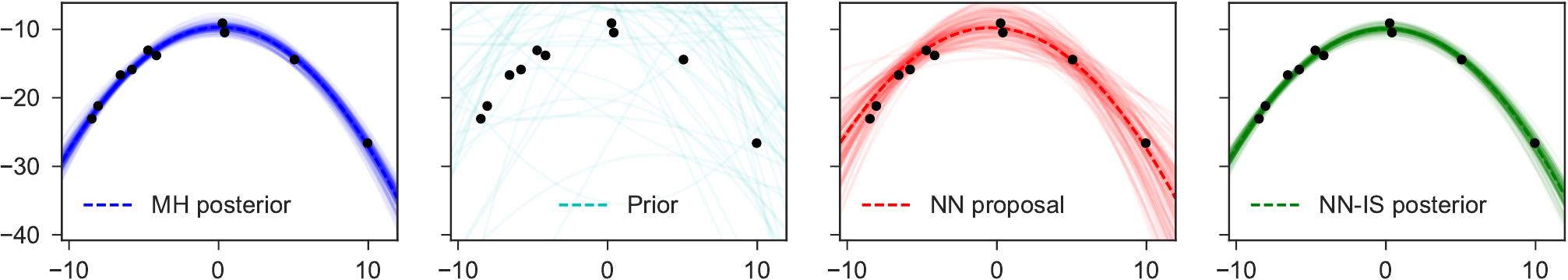}\vspace{0.5em}\\
	\includegraphics[height=1.05in]{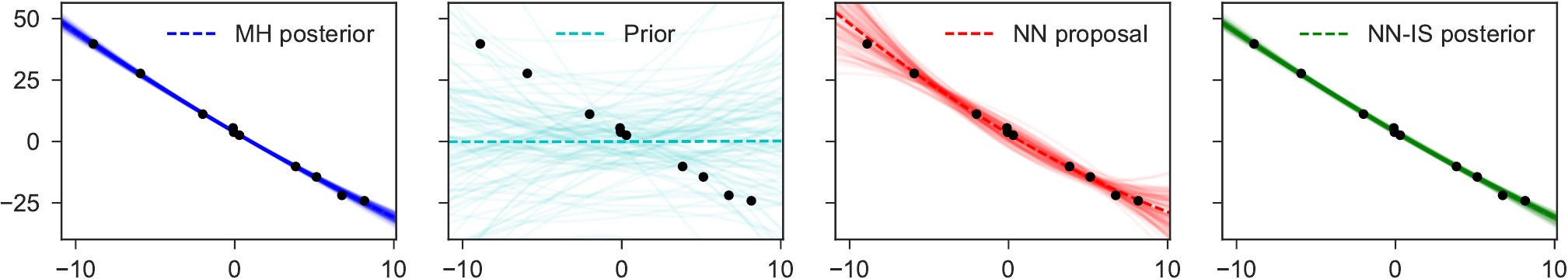}\vspace{0.5em}\\
	\includegraphics[height=1.05in]{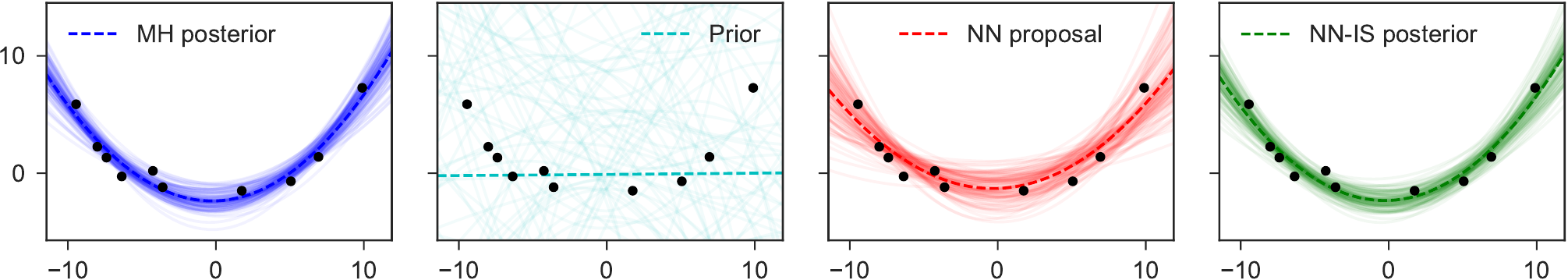}\vspace{0.5em}\\
	\includegraphics[height=1.05in]{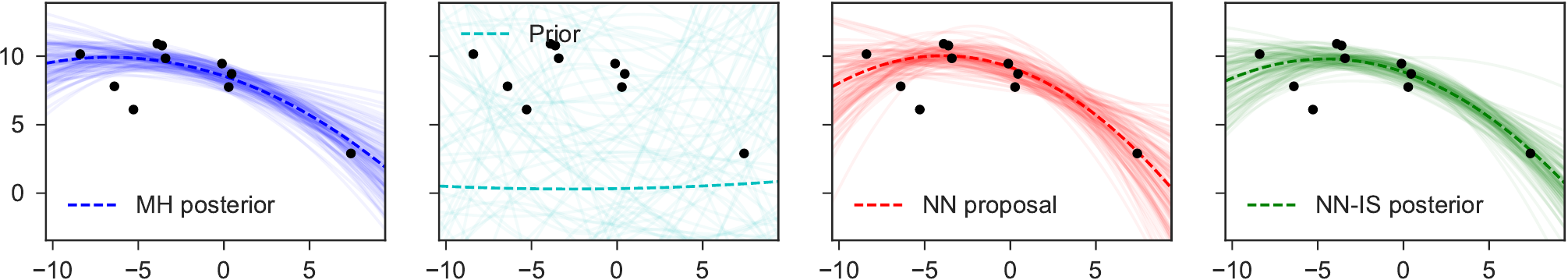} \vspace{0.5em}\\
	\ ~\includegraphics[height=1.05in]{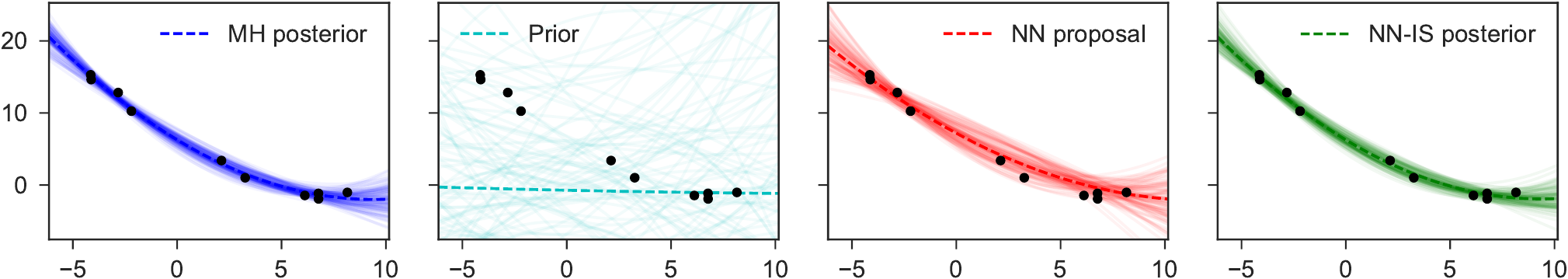}
\vspace{-.5em}
\caption{Representative output in the polynomial regression example.
	Plots show 100 samples each at 5\% opacity, with the mean marked as a solid dashed line.
	These are all proposed using the same pre-trained neural network --- not just the same neural network structure,
	but also identical learned weights.
	The MCMC posterior is generated by thinning 10000 samples by a factor 100, after 10000 samples of burnin.
	The neural network proposal yields estimated polynomial curves close to the true posterior 
	solution, albeit slightly more diffuse.
	}
\vspace{-0.3em}
\label{fig:regression-output}
\end{figure*}

\section{Examples}

\subsection{Inverting a single factor}

To illustrate the basic method for inverting factors, we consider a 
non-conjugate polynomial regression model, 
with global-only latent variables.
The graphical model, its inversion, and the neural network structure are shown in Figure~\ref{fig:regression}.
Here we place a Laplace prior on the regression weights, and have Student-t likelihoods,
giving us
\begin{align}
w_d &\sim \mathrm{Laplace}(0, 10^{1-d}) &\text{ for } d &= 0,1,2; \nonumber \\
t_n &\sim \mathrm{t}_\nu(w_0 + w_1z_n + w_2z_n^2, \epsilon^2) &\text{ for } n &= 1,\dots,N \nonumber
\end{align}
for fixed $\nu=4, \epsilon=1$, and $z_n \in (-10,10)$ uniformly.
The goal is to estimate the posterior distribution of weights for the constant, linear, and quadratic terms,
given any possible collected dataset $\{ z_n, t_n \}_{n=1}^N$.
In the notation of the preceding sections, we have latent variables $\+x \equiv \{ w_0, w_1, w_2 \}$ 
and observed variables $\+y \equiv \{ z_n, t_n \}_{n=1}^N$.

Note particularly that although the original graphical model which expressed $p(\+y|\+x)p(\+x)$
factorizes into products over $y_n$ which are conditionally independent given $\+x$,
in the inverse model $\tilde p(\+x|\+y)$ due to the explaining-away phenomenon all
latent variables depend on all others:
there are no latent variables which can be $d$-separated from the observed $\+y$,
and all latent variables share $\+y$ as parents.
This means we fit as proposal only a single joint density $q(w_{0:2} | z_{1:N}, t_{1:N})$.
Examples of representative output from this network are shown in Figure~\ref{fig:regression-output}.
The trained network used here 300 hidden units in each of two hidden layers, and a mixture of 3 Gaussians as each output.

\subsection{A hierarchical Bayesian model}

Consider as a new example a representative multilevel model where exact inference is intractable, 
a Poisson model for estimating failure rates of power plant pumps \citep{george1993conjugate}.
Given $N$ power plant pumps, each having operated for $t_n$ thousands of hours,
we see $x_n$ failures, following
\begin{align}
\alpha &\sim \mathrm{Exponential}(1.0), &
\beta &\sim \mathrm{Gamma}(0.1, 1.0), \nonumber \\
\theta_n &\sim \mathrm{Gamma}(\alpha, \beta), &
y_n &\sim \mathrm{Poisson}(\theta_nt_n). \nonumber
\end{align}
The graphical model,
an inverse factorization,
and the neural network structure are shown in Figure~\ref{fig:poisson}.
To generating synthetic training data, $t_n$ are sampled {\em iid}~from an exponential distribution with mean 50.

The repeated structure in the inverse factorization of this model allows us to learn a single inverse factor 
to represent the distribution $\tilde p(\theta_n|t_n, y_n)$ across all $n$.
This yields a far simpler learning problem than were we forced to fit all of $\tilde p(\theta_{1:N}|t_{1:N}, y_{1:N})$
jointly.
Further, the repeated structure allows us to use a divide-and-conquer SMC algorithm \cite{lindsten2014divide}
which works particularly efficiently on this model.
Each of the $N$ replicated structures are sampled in parallel with independent particle sets, weighted locally, and resampled;
once all $\theta_n$ are sampled, we end by sampling $\alpha$ and $\beta$ jointly, 
which need both be included in order to evaluate the final terms in the joint target density.
We stress that there is no obvious baseline proposal density to use for a divide-and-conquer SMC algorithm,
as neither the marginal prior nor posterior distributions over $\theta_n$ are available in closed form.
Any usage of this algorithm requires manual specification of some proposal $q(\theta_n)$.

\begin{figure}[t]
\centering
\includegraphics[height=0.19\textwidth]{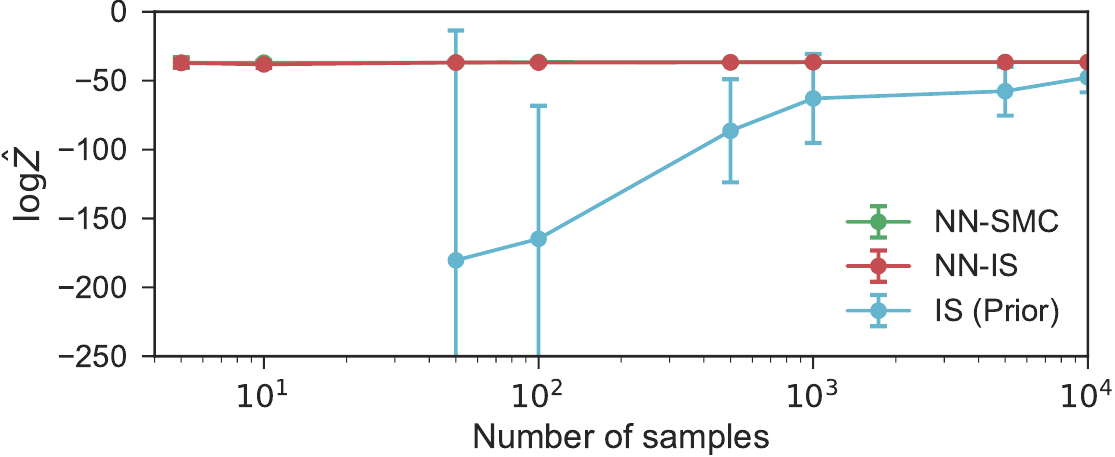}
\vspace{-0.5em}
\caption{Convergence of marginal likelihood estimate as a function of number of particles,
for likelihood-weighted importance sampling, neural network importance sampling, and
a divide-and-conquer sequential Monte Carlo algorithm with neural network proposals.
The SMC algorithm can achieve reasonable estimates of the normalizing constant with as few as 5 samples.
Plot shows mean of 10 runs; error bars show two standard deviations.
}
\label{fig:poisson-ml}
\end{figure}

We test our proposals on the actual power pump failure data analyzed in \citet{george1993conjugate}.
The relative convergence speeds of marginal likelihood estimators from importance sampling from prior and neural network proposals, 
and SMC with neural network proposals, are shown in Figure~\ref{fig:poisson-ml}.
To capture the wide tails of the broad gamma distributions, we use a mixture of 10 Gaussians here at each output node, and 500 hidden units in each of two hidden layers.

\subsection{Factorial hidden Markov model}

Proposals can also be learned to approximate the optimal
filtering distribution in models for sequential data; we demonstrate here
on a factorial hidden Markov model \citep{ghahramani1997factorial},
where each time step has a combinatorial latent space.
The additive model we consider is inspired by the model studied in \citet{kolter2012approximate}
for disaggregation of household energy usage;
effective inference in this model is a subject of continued research.
Some number of devices $D$ are either in an active state,
in which case each device $i$ consumes $\mu^i$ units of energy,
or it is off, in which case it consumes no energy.
At each time step we receive a noisy observation of the total amount of energy consumed,
summed across all devices.
This model, whose graphical model structure is shown in Figure~\ref{fig:factorial-hmm}, can be represented as
\begin{align*}
x^i_t|x^i_{t-1} &\sim \mathrm{Bernoulli}( \theta^i[x^i_{t-1}] ) \\
y_t  | x^1_t,\dots,x^D_t &\sim \Normal\big( \textstyle\sum_{i=1}^D \mu^i x^i_t, \sigma^2 \big),
\end{align*}
where $\theta^i$ represents the prior probability of devices switching on or off at each time increment.
We design a synthetic example with $D=20$, meaning each time step has $2^{20} \approx 100,000$ possible discrete states;
the parameters $\mu^{d}$ are spread out from 30 to 500, with $\sigma = 20$.
Each individual device has an initial probability $0.1$ of being activated at $t=1$, switching state at subsequent $t$ with probability $0.05$.

\begin{figure}[t]
\centering
\includegraphics[height=0.19\textwidth]{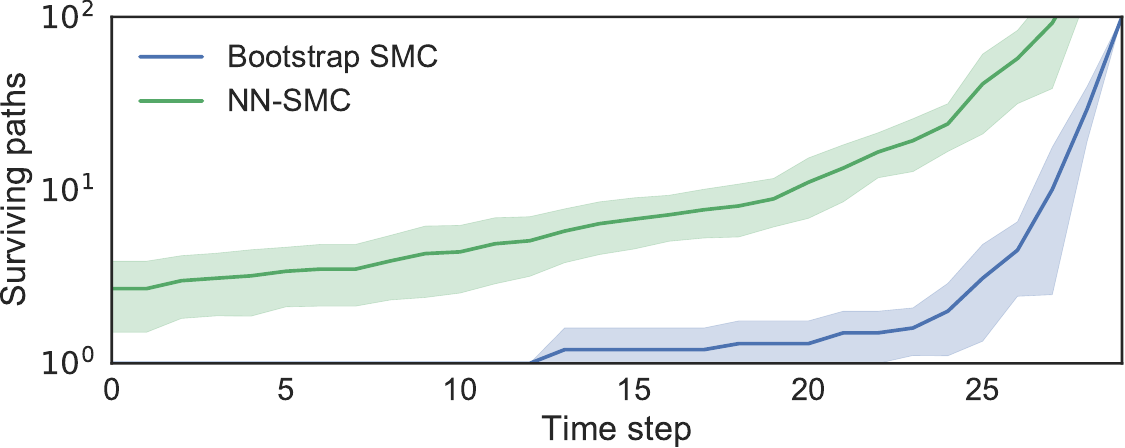}
\vspace{-1.7em}
\caption{Learned proposals reduce particle degeneracy in the factorial HMM.
Here we show the number of unique ancestries which survive over the course of 30 time steps, running 100 particles.
Proposing from the transition dynamics nearly immediately degenerates to a single possible solution;
the learned proposals increase the effective sample size at each stage and reduce the need for resampling.
Plot shows mean and standard deviation over 10 runs.
}
\vspace{1.15em}
\label{fig:hmm-ess}
\end{figure}

As different combinations of devices can yield identical total energy usage
it is impossible to disambiguate between different combinations of active devices from a single observation,
meaning any successful inference algorithm must attempt to mix across many disconnected modes over time
to preserve the multiple possible explanations.
The effect of the learned proposals on the overall number of surviving particles is shown in Figure~\ref{fig:hmm-ess}.
Our proposal model uses $D$ Bernoulli outputs in a 4-layer network, with 300 units per hidden layer;
it takes as input the $D$ latent states at the previous time $t-1$, as well as the current observation $y_t$.

\section{Discussion}

We present this work primarily as a manner by which  
we compile away application-time inference costs when performing SMC,
and automating the manual task of designing proposal densities.
However, in some situations
direct sampling from the model may provide a satisfactory approximation even eschewing importance weighting steps;
in such cases our approach can be viewed as a graphical-model-regularized algorithm for designing and training neural networks with interpretable structural representations.
Rather than learning from data, the emulator model is chosen to approximate the specified generative model, akin to the ``sleep'' cycle of the wake-sleep algorithm \cite{hinton1995wake}.

In contrast to variational autoencoders \citep{kingma2014auto}, where one
simultaneously learns parameters for both the inference network and generative model from data,
we assume a known generative model with fixed parameters and structured, interpretable latent variables.
This provides robustness to bias arising from training data which comes from an unrepresentative sample, and also allows us to apply our method in situations where a sufficiently large supply of exemplar data is unavailable.
However, it does require placing trust in the generative model: in particular, it requires a generative model which could plausibly create the data we will later collect and condition on.

Beyond these differences, our choice of $D_{KL}(\pi||q)$, the same minimized by EP, leads to approximations more appropriate for SMC refinement than a variational Bayes objective function;
see e.g.~\citet{minka2005divergence} for a discussion of ``zero-forcing'' behavior, and e.g.~\citet{cappe2008adaptive} for a discussion of pathological cases in learned importance sampling distributions.

\section*{Acknowledgements} 
BP would like to thank both Tom Jin and Jan-Willem van de Meent for their helpful discussions, feedback, and ongoing commiseration.
FW is supported under DARPA PPAML through the U.S. AFRL under Cooperative Agreement number FA8750-14-2-0006, Sub Award number 61160290-111668.

{\normalsize
\bibliography{../refs.bib}
\bibliographystyle{apalike}
}

\end{document}